\documentclass[letterpaper]{article} 
\usepackage{aaai2027}  
\nocopyright
\usepackage[hyphens]{url}  
\usepackage{graphicx} 
\usepackage{natbib}  
\usepackage{caption} 
\usepackage{algorithm}
\usepackage{algorithmic}
\usepackage{amsmath}
\usepackage{array}
\usepackage{tabularx}
\usepackage{xspace}
\newcolumntype{C}{>{\centering\arraybackslash}c}
\newcommand{\methodname}{\textsc{PCSD}\xspace}
\usepackage{soul}
\usepackage[table]{xcolor} 
\definecolor{topcolor}{RGB}{252, 236, 196}
\definecolor{secondcolor}{RGB}{223, 235, 253}
\usepackage{newfloat}
\usepackage{listings}
\DeclareCaptionStyle{ruled}{labelfont=normalfont,labelsep=colon,strut=off} 
\floatstyle{ruled}
\newfloat{listing}{tb}{lst}{}
\floatname{listing}{Listing}

\usepackage{booktabs}

\title{PCSD: Persistent Consistency for Self-Distillation in Agentic  \\ Reinforcement Learning}
\author{
    Chunji Lv\textsuperscript{\rm 1,2},
    Yangguang Wei\textsuperscript{\rm 2},
    Junlin Liu\textsuperscript{\rm 3},
    Yang Gao\textsuperscript{\rm 2},
    Ming Liu\textsuperscript{\rm 2}, 
    Xinming Wang\textsuperscript{\rm 3},
    Jinyang Wu\textsuperscript{\rm 4},\\
    Guoren Wang\textsuperscript{\rm 1},
    Changsheng Li\textsuperscript{\rm 1}\thanks{Corresponding author.}
}

\affiliations{
    \textsuperscript{\rm 1}Beijing Institute of Technology\\
    \textsuperscript{\rm 2}Meituan\\
    \textsuperscript{\rm 3}Institute of Automation, Chinese Academy of Sciences\\
    \textsuperscript{\rm 4}Tsinghua University\\
    3120250994@bit.edu.cn
}

\begin{document}

\maketitle

\begin{abstract} 
Large language model agents have shown strong potential in complex interactive tasks, yet their reinforcement learning (RL) is often hindered by sparse rewards, as a long multi-turn trajectory may receive only a single outcome-level signal. On-policy self-distillation (OPSD) provides dense token-level supervision from a privileged teacher, but the teacher may not be reliable at every position. Existing methods commonly rely on isolated token-level discrepancies, which can be sensitive to noise, or assign a shared step-level weight that may overlook positional variation. We propose Persistent Consistency Self-Distillation (PCSD), which derives token-level distillation weights from the local persistence of teacher-favoring signals. PCSD combines adaptive windows with exponentially decayed aggregation to capture persistent relative teacher support, applies trend-aware modulation to attenuate locally declining support, and produces continuous weights through sigmoid gating. The resulting objective is jointly optimized with GRPO, combining dense teacher guidance with sparse environmental feedback. Without inference-time skills, PCSD achieves the best ALFWorld Overall results among all baselines on both backbones, exceeding GRPO by 15.6 and 13.3 points and SDAR by 6.2 and 5.5 points, while remaining competitive on WebShop and gaining 15.8 points over GRPO on unseen ALFWorld split. \end{abstract}

\begin{figure*}[t]

    \centering
    \includegraphics[width=0.85\linewidth]{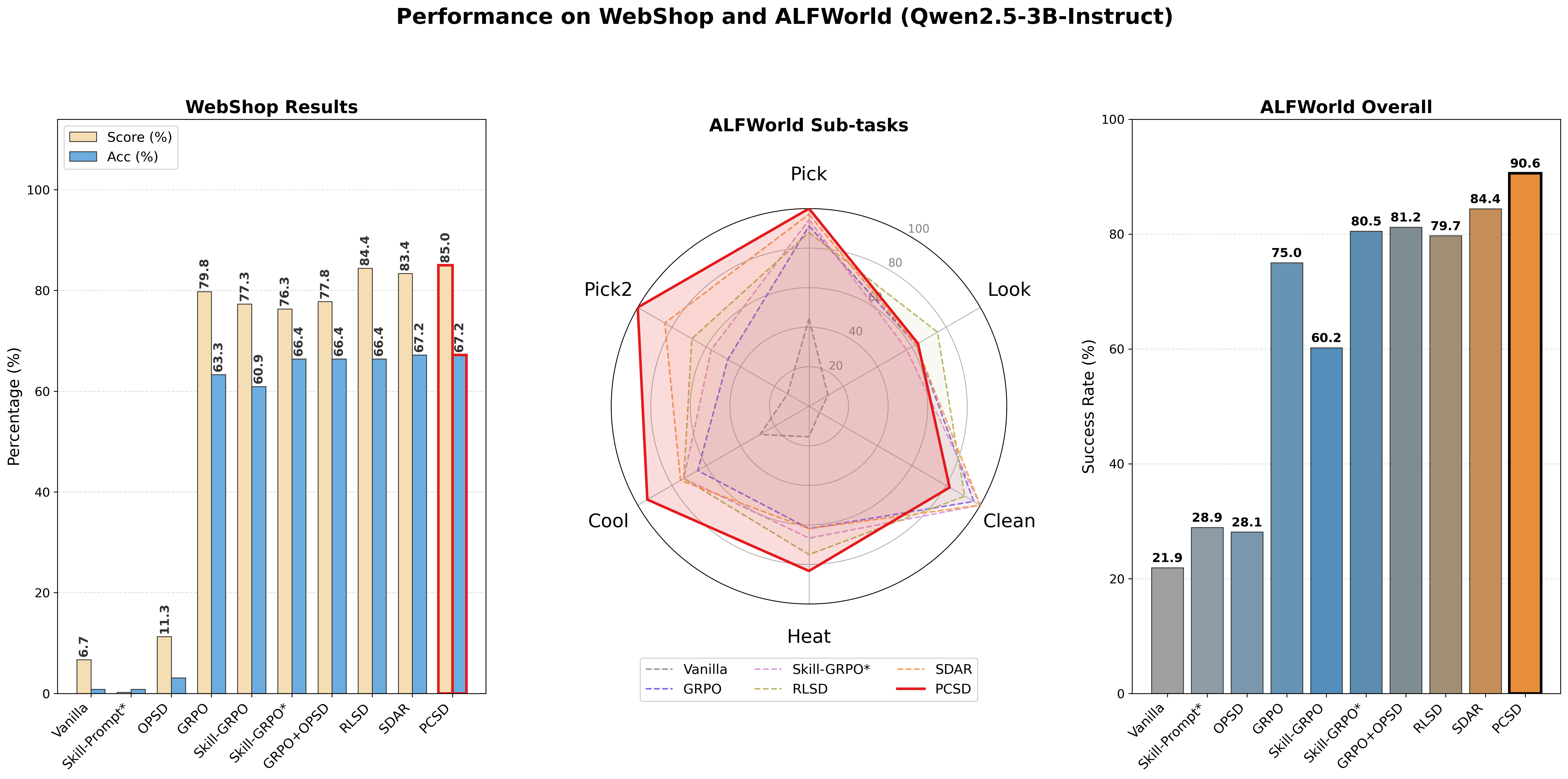}
    \caption{Overall performance comparison. Main results on WebShop and ALFWorld with Qwen2.5-3B-Instruct. Left: WebShop Score and Acc; Middle: ALFWorld sub-task radar; Right: ALFWorld Overall success rate.}
    \label{fig:overall_performance}
    \vspace{-7pt}
\end{figure*}

\section{Introduction}

Large language models (LLMs) have shown strong potential as autonomous agents for complex interactive tasks~\cite{jin2025search,li2026webthinker,qian2026toolrl,wang2025hitchhiker,shridhar2020alfworld,yao2022webshop,shinn2023reflexion,acharya2025agentic}. Unlike static single-turn inference, LLM agents must make a sequence of interdependent decisions over extended multi-turn interactions, where each action changes the subsequent environmental state and may affect the final outcome many turns later. Reinforcement learning (RL) methods such as Group Relative Policy Optimization (GRPO) provide a natural framework for optimizing such sequential behavior~\cite{deepseekai2026deepseekv4}. However, their effectiveness is often constrained by reward sparsity: a trajectory spanning dozens of turns and hundreds of generated tokens may receive only a single scalar reward at its end. Such sparse and delayed feedback provides little guidance on which decisions contribute to success or failure, making credit assignment particularly challenging.

On-policy self-distillation (OPSD)~\cite{zhao2026self} alleviates reward sparsity by using a privileged teacher---typically a frozen copy of the student augmented with additional context---to provide dense token-level supervision on trajectories sampled from the current policy. However, privileged information does not guarantee reliable guidance at every position: imperfect retrieval, noisy context, and task ambiguity may cause the teacher to favor suboptimal tokens. Indiscriminate distillation can therefore transfer erroneous preferences, making it crucial to identify \emph{which token positions contain trustworthy teacher signals}.

Existing credibility-aware distillation methods typically estimate teacher reliability at either the token or step level. Token-level methods~\cite{lu2026self,wang2026tcod,yang2026self} compute a distillation weight from an isolated teacher--student discrepancy or probability ratio at each position. While preserving fine-grained positional information, such pointwise estimates are sensitive to sampling variability: a large discrepancy at a single token may indicate either meaningful teacher advantage or a transient fluctuation. Step-level methods~\cite{zhong2026sod} instead aggregate discrepancy signals over an entire interaction or reasoning step and assign a shared weight to all tokens within that step. Although this aggregation improves robustness to local noise, it may obscure substantial variation in teacher credibility across positions. Existing approaches therefore face a fundamental trade-off between token-level precision and aggregation-based robustness.

To address this trade-off, we propose \textbf{Persistent Consistency Self-Distillation (PCSD)}, a token-level weighting framework that estimates teacher credibility from the local persistence of teacher-favoring signals. Our key insight is that informative teacher advantage should persist across a local neighborhood, whereas isolated spikes are more likely to reflect sampling noise or incidental variation. PCSD therefore evaluates each token using both its pointwise discrepancy and the persistence of supporting evidence at nearby positions, combining token-level resolution with the robustness of local aggregation.

PCSD implements this principle through three complementary mechanisms. First, it adaptively adjusts the aggregation window according to the local statistical properties of teacher--student discrepancies, using a broader range in noisy regions and a finer range in stable regions. Second, it aggregates nearby discrepancy signals with exponential decay, assigning greater importance to positions closer to the current token and thereby preserving positional locality. Third, PCSD applies trend-aware modulation to attenuate declining teacher support, reducing reliance on transient signals, and maps the resulting credibility to a token-level weight via sigmoid gating. The PCSD-weighted distillation objective is jointly optimized with GRPO, allowing dense teacher supervision to complement sparse environmental feedback.

We evaluate PCSD on ALFWorld and WebShop across two model backbones. As shown in Figure~\ref{fig:overall_performance}, PCSD consistently outperforms outcome-only GRPO and GRPO with existing self-distillation weighting schemes, while generalizing robustly to unseen scenarios. These results demonstrate that persistent local evidence effectively identifies credible teacher signals while filtering unreliable guidance.

Our contributions are summarized as follows:

\begin{itemize}
    \item We propose PCSD, a token-level on-policy self-distillation framework that estimates teacher credibility from the local persistence of teacher-favoring signals, combining fine-grained positional discrimination with robustness to pointwise noise.

    \item We develop an adaptive aggregation mechanism that adjusts its effective range according to the local statistics of teacher--student discrepancies and uses exponential decay to preserve positional locality.

    \item We introduce trend-aware modulation and continuous sigmoid gating to attenuate locally declining or unstable teacher support and transform credibility estimates into smooth token-level distillation weights.

    \item Extensive experiments show that PCSD substantially outperforms outcome-only GRPO and existing GRPO-based weighting methods on ALFWorld across both model scales, remains competitive on WebShop, and generalizes robustly to unseen scenarios.
\end{itemize}

\section{Related Work}
\subsection{Agentic Reinforcement Learning.} 
Reinforcement learning (RL)~\cite{schulman2017proximal,rafailov2023direct} has become a dominant paradigm for LLM post-training~\cite{singh2025openai,comanici2025gemini,guo2025deepseek,yang2025qwen3,liu2025deepseek,team2025kimi,team2026kimi,zeng2025glm,team2025longcat}, and recent work has extended it to agentic interaction trajectories, including code generation~\cite{jimenez2024swe,gehring2024rlef}, tool use~\cite{yao2022react,yao2024tau,lv2026physagent,jin2025search,mallen2023not}, GUI interaction~\cite{rawles2025androidworld,ye2025mobile}, and web navigation~\cite{qi2025webrl,shridhar2020alfworld,yao2022webshop}. A central challenge in these long-horizon settings is learning from sparse and delayed feedback. Environments typically provide only trajectory-level outcome signals, making it difficult to identify which intermediate actions or tokens are responsible for the final result. Critic-free methods~\cite{shao2024deepseekmath,guo2025deepseek,feng2026group,yu2026dapo,tan2025gtpo,zhang2026gvpo} improve the scalability of RL by replacing explicit value-function learning with group-relative rewards. However, their supervision remains coarse-grained. Our work addresses this limitation by incorporating on-policy self-distillation (OPSD) to provide dense token-level supervision. While preserving the RL optimization framework, OPSD complements sparse environmental feedback with dense teacher guidance, thereby improving fine-grained credit assignment in long-horizon agentic tasks.

\subsection{On-Policy Self-Distillation.}
In agentic RL, on-policy distillation (OPD)~\cite{agarwal2024policy,yang2025qwen3,deepseekai2026deepseekv4,ye2026policy,li2026rethinking,ye2025black,wang2026mad} complements sparse trajectory-level rewards with dense teacher guidance on policy-sampled trajectories, improving fine-grained credit assignment. On-policy self-distillation (OPSD)~\cite{zhao2026self,hubotter2026reinforcement,yang2024self} realizes this idea by using a frozen, privileged-context-augmented copy of the student as the teacher, avoiding the cost of training a separate teacher model. Nevertheless, reliably assessing teacher guidance at the token level remains challenging. Existing methods are either token-level~\cite{lu2026self,yang2026self,yang2026opid,wang2026tcod,wu2026seed,lu2026skill0}, using token-wise teacher--student divergence or probability ratios to estimate reliability from a single observation, or step-level~\cite{zhong2026sod,zhang2026stepopsd}, aggregating divergence over an entire reasoning or action step and assigning uniform weights to all tokens within it. The former is sensitive to local noise and may confuse sampling fluctuations with true distributional disagreement, while the latter is too coarse to capture positional variations in teacher reliability. Our work addresses this limitation with persistent consistency, which determines teacher-signal usability by testing whether teacher-favoring support persists over a local window, thereby adaptively balancing token-level precision and step-level robustness.

\begin{figure*}[t]

\centering
\includegraphics[width=0.9\textwidth]{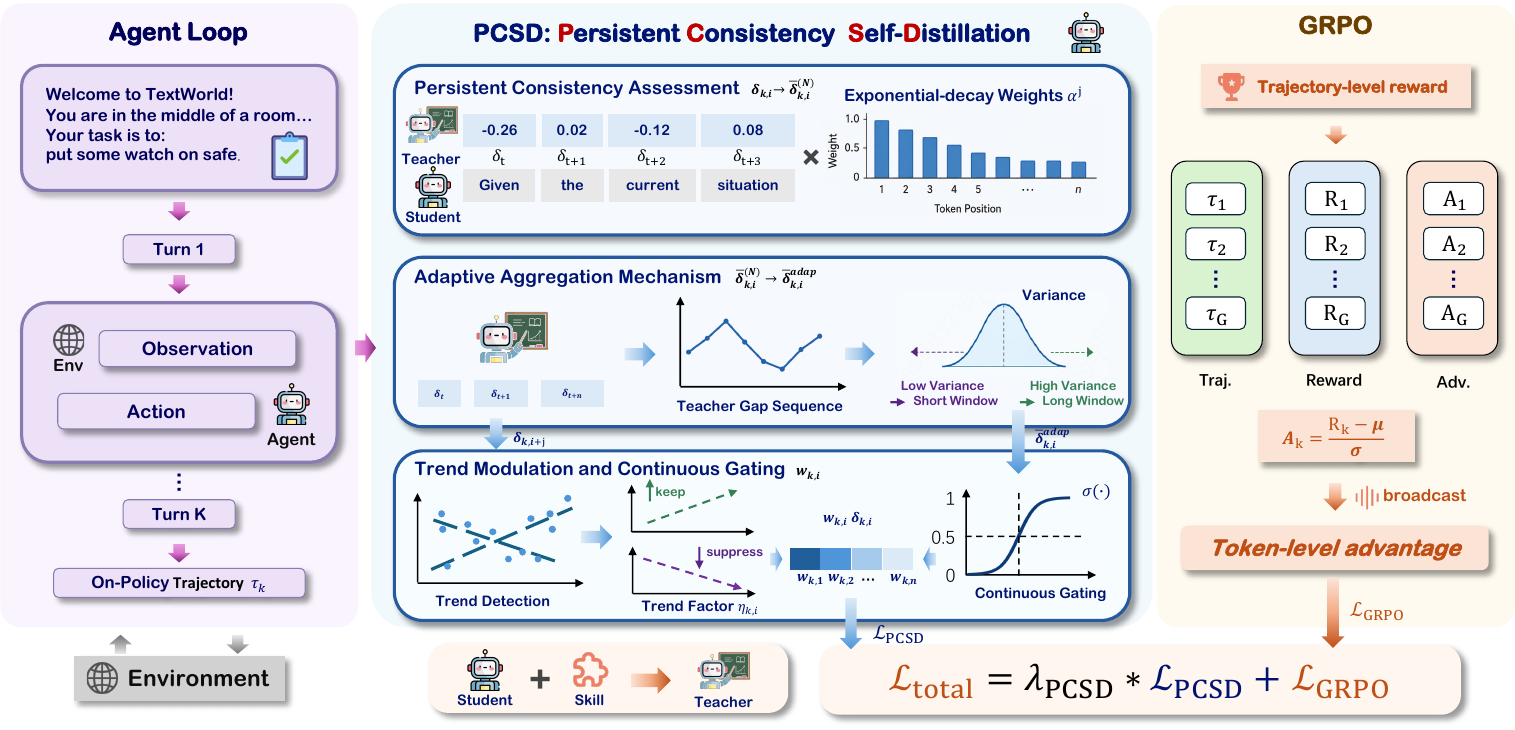} 
\caption{Framework of PCSD. The student collects on-policy trajectories via multi-turn interaction, while a frozen, skill-augmented teacher scores the student-generated tokens. PCSD aggregates teacher--student log-probability gaps across multiple time scales with exponential decay, adaptively weights them by local variability, and applies trend modulation and continuous gating to derive token-level distillation weights, which are jointly optimized with the trajectory-reward-based GRPO objective.}
\label{fig2}
\vspace{-7pt}
\end{figure*}


\section{Preliminaries}

\subsection{Multi-turn Agent Reinforcement Learning}
We consider a multi-turn agent task in which an LLM agent interacts with an external environment over multiple turns. At turn $k$, the environment provides a state $s_k$, and the student policy $\pi_\theta$ generates a response ${y}_k$:
\begin{equation}
\mathbf{y}_k = (y_{k,1}, y_{k,2}, \ldots, y_{k,T_k})
\sim \pi_\theta(\cdot \mid s_k),
\end{equation}
where $T_k$ denotes the response length. The generated response may trigger an action in the environment, leading to the next state $s_{k+1}$. A complete trajectory can be written as
\begin{equation}
\tau = (s_1, \mathbf{y}_1, s_2, \mathbf{y}_2, \ldots, s_k, \mathbf{y}_k),
\end{equation}
where $k$ is the number of interaction turns. In agentic RL, environmental feedback is often sparse and delayed: the reward is typically available only at the trajectory level,  e.g., a terminal success or failure signal. This makes it difficult to assign credit to individual intermediate tokens or actions.

\subsection{On-Policy Self-Distillation}
On-policy self-distillation (OPSD) augments sparse environmental feedback
with dense teacher guidance on trajectories sampled from the current student
policy. Given an on-policy response $\mathbf{y}_k$, the student predicts each
token based on its visible history, while the teacher additionally conditions
on privileged context. Specifically, for token position $i$, we denote the
student-visible context by $h_{k,i}^{S}$ and the teacher-augmented context by
$h_{k,i}^{T}$, where $h_{k,i}^{T}$ includes task-relevant skills unavailable
to the student.

The privileged-context teacher achieves substantially higher trajectory-level success rates than the student, as empirically established in prior OPSD studies~\cite{zhao2026self}. We therefore treat the teacher as a stronger source of supervision on average. However, this trajectory-level advantage does not imply that its guidance is equally useful at every token position. The relevance of the retrieved skills and the teacher's support for student-sampled tokens may vary along a response. The purpose of \methodname is therefore not to determine whether the teacher is globally stronger, but to allocate distillation strength according to the local persistence of its token-level support.

Let $\pi_T$ denote the frozen privileged-context teacher policy, initialized from the same base checkpoint as the student and kept fixed throughout training. The teacher--student sampled log-probability gap is defined as
\begin{equation}
\label{eq:token_gap}
\delta_{k,i}
=
\log \pi_T(y_{k,i} \mid h_{k,i}^{T})
-
\log \pi_\theta(y_{k,i} \mid h_{k,i}^{S}).
\end{equation}
We treat $\delta_{k,i}$ as a continuous measure of the teacher's relative support for the sampled token $y_{k,i}$. Larger values indicate stronger relative support. \methodname uses its local persistence to form continuous token-level distillation weights.

\section{Method}
\label{sec:method}

In this section, we introduce \methodname (Persistent Consistency Self-Distillation), an on-policy self-distillation method that selectively allocates teacher supervision across token positions. As described in the preliminaries, the privileged-context teacher is substantially stronger than the student at the trajectory level. Nevertheless, the usefulness of its guidance can vary locally across sampled tokens.

\methodname computes continuous token-level distillation weights from persistent patterns in the teacher--student sampled log-probability gap. We operationalize persistent consistency as sustained teacher support for student-sampled tokens over nearby positions. Starting from the token-level gap sequence $\{\delta_{k,i}\}$, \methodname first constructs an exponentially weighted persistent-consistency estimate and then adaptively interpolates between short- and long-window estimates according to local gap variability. It next applies one-sided trend modulation and sigmoid gating to produce continuous token-level distillation weights. Finally, the weights allocate auxiliary supervision across student-sampled tokens, and the resulting objective is jointly optimized with GRPO.

\subsection{Persistent Consistency Assessment}
\label{sec:persistent_consistency}

We first estimate the local usability of teacher guidance by aggregating teacher--student sampled log-probability gaps over a local window. Given the token-level gap $\delta_{k,i}$, a larger value indicates that the teacher assigns a higher probability relative to the student to the sampled token $y_{k,i}$. We interpret this quantity as the teacher's relative support for reinforcing the sampled token. 

A pointwise gap provides evidence from only one token position and may vary substantially across adjacent tokens. Such variation can arise from sampling fluctuations as well as meaningful positional differences in the teacher--student distributions. Consequently, a single-position observation may provide an unstable basis for allocating distillation strength.

To incorporate local evidence, we aggregate gaps over a forward window of size $N$ with exponential weighting:
\begin{equation}
\bar{\delta}_{k,i}^{(N)}
=
\frac{
\sum_{j=0}^{N-1}
\alpha^j m_{k,i+j} \delta_{k,i+j}
}{
\sum_{j=0}^{N-1}
\alpha^j m_{k,i+j}
},
\end{equation}
where $\alpha \in (0,1)$ is the decay factor and $m_{k,i}$ is the response
mask. For positions near the end of a response, the window is truncated and the aggregation is normalized over valid tokens only. This formulation assigns larger weights to positions closer to the current token while still incorporating evidence from the surrounding local region. The forward-looking window is computed after the complete response is sampled and is used only to construct the training objective.

We use $\bar{\delta}_{k,i}^{(N)}$ as the persistent-consistency estimate,
which summarizes the teacher's relative support across the local window
rather than relying solely on the pointwise gap at the current position.
Larger values indicate stronger persistent support for the sampled tokens.
By incorporating evidence from neighboring positions, this estimate is less
dependent on an isolated token-level observation while retaining a
position-specific value for subsequent weighting.

\subsection{Adaptive Aggregation Mechanism}
\label{sec:adaptive_aggregation}

A single fixed window may not adequately capture heterogeneous gap patterns
across a response. A short window preserves fine-grained positional
information but is more sensitive to local fluctuations, whereas a long
window provides stronger smoothing at the cost of reduced positional
resolution. We therefore implement soft adaptive windowing by adjusting the
relative contributions of short- and long-window persistent-consistency
estimates according to the local variability of the gap sequence.

For token position $i$ at turn $k$, we first compute the local mean over a
forward-looking maximum analysis window of size $N_{\max}$:
\begin{equation}
\mu_{k,i}^{(N_{\max})}
=
\frac{
\sum_{j=0}^{N_{\max}-1}
m_{k,i+j}\delta_{k,i+j}
}{
\sum_{j=0}^{N_{\max}-1}
m_{k,i+j}
},
\end{equation}
and the corresponding local variance:
\begin{equation}
\sigma^2_{k,i}
=
\frac{
\sum_{j=0}^{N_{\max}-1}
m_{k,i+j}
\left(
\delta_{k,i+j}
-
\mu_{k,i}^{(N_{\max})}
\right)^2
}{
\sum_{j=0}^{N_{\max}-1}
m_{k,i+j}
}.
\end{equation}
Here, $m_{k,i}$ is the response mask. 

We then map the local variance to a normalized interpolation coefficient:
\begin{equation}
r_{k,i}
=
\operatorname{clip}
\left(
\frac{
\sigma^2_{k,i}
-
\tau_{\mathrm{low}}
}{
\tau_{\mathrm{high}}
-
\tau_{\mathrm{low}}
},
0,1
\right),
\end{equation}
where $\tau_{\mathrm{low}}$ and $\tau_{\mathrm{high}}$ are variance thresholds satisfying $\tau_{\mathrm{low}}<\tau_{\mathrm{high}}$. The coefficient $r_{k,i}$ controls the aggregation scale: a larger value assigns more weight to the long-window estimate, whereas a smaller value favors the short-window estimate. 

Specifically, we define the adaptive persistent-consistency estimate as
\begin{equation}
\bar{\delta}_{k,i}^{\mathrm{adaptive}}
=
(1-r_{k,i})
\bar{\delta}_{k,i}^{(N_{\min})}
+
r_{k,i}
\bar{\delta}_{k,i}^{(N_{\max})}.
\end{equation}
When the local gap variance is below $\tau_{\mathrm{low}}$, the mechanism
reduces to the short-window estimate
$\bar{\delta}_{k,i}^{(N_{\min})}$, preserving greater token-level
resolution. When the variance exceeds $\tau_{\mathrm{high}}$, it reduces to
the long-window estimate
$\bar{\delta}_{k,i}^{(N_{\max})}$, providing stronger local smoothing.
Intermediate variance values yield a continuous interpolation between these
two aggregation scales. This soft adaptive mechanism allows \methodname to
adjust the smoothing strength across token positions without requiring
discrete, position-specific window operations.

\subsection{Trend Modulation and Continuous Gating}
\label{sec:denoising_gating}

Exponential-decay aggregation places greater emphasis on nearby positions, but a large current gap followed by rapidly decreasing teacher support may still dominate the local estimate. To address this pattern, we apply one-sided trend modulation before computing the final distillation weight.

To estimate this trend, we use the maximum window $N_{\max}$ at every position, providing a common analysis scale with more observations and less sensitivity to individual token-level variations.

For token position $i$ at turn $k$, we compute a mask-aware ordinary least
squares (OLS) slope over the maximum analysis window. We first calculate the
mean valid position:
\begin{equation}
\bar{j}_{k,i}
=
\frac{
\sum_{j=0}^{N_{\max}-1} m_{k,i+j}j
}{
\sum_{j=0}^{N_{\max}-1} m_{k,i+j}
},
\end{equation}
and then estimate the local slope:

\begin{equation}
\mathrm{slope}_{k,i}
=
\frac{
\sum_{j=0}^{N_{\max}-1}
m_{k,i+j}(j-\bar{j}_{k,i})\delta_{k,i+j}
}{
\sum_{j=0}^{N_{\max}-1}
m_{k,i+j}(j-\bar{j}_{k,i})^2
+\epsilon_{\mathrm{slope}}
}.
\end{equation}
where $\epsilon_{\mathrm{slope}}>0$ ensures numerical stability and defaults the slope to zero when fewer than two valid tokens exist.

We apply a one-sided modulation based on the local slope. Positions with negative slopes, which indicate decreasing relative teacher support, are attenuated, whereas nonnegative slopes receive no additional trend-based attenuation. This asymmetric design complements exponential decay: the trend factor attenuates declining local support, while exponential weighting limits distant-position contributions.

Specifically, the trend modulation factor is
\begin{equation}
\eta_{k,i}
=
\operatorname{clip}
\left(
1
-
\gamma \cdot
\operatorname{ReLU}
\left(
-\frac{\mathrm{slope}_{k,i}}
{\delta_{\mathrm{scale},k}}
\right),
0,1
\right),
\end{equation}
where $\gamma$ controls the modulation strength. We normalize the slope using
the mean absolute gap within the response $\mathbf{y}_k$:
\begin{equation}
\delta_{\mathrm{scale},k}
=
\frac{
\sum_i m_{k,i}|\delta_{k,i}|
}{
\sum_i m_{k,i}
}
+
\epsilon_{\mathrm{scale}},
\end{equation}
where $\epsilon_{\mathrm{scale}}>0$ prevents division by a near-zero scale. 

Finally, we map the adaptive persistent-consistency estimate to a bounded
continuous token-level distillation weight:
\begin{equation}
w_{k,i}
=
\sigma
\left(
\beta_{\mathrm{gate}} \cdot \bar{\delta}_{k,i}^{\mathrm{adaptive}}
\right)
\cdot
\eta_{k,i},
\end{equation}
where $\beta_{\mathrm{gate}}$ controls the sharpness of the sigmoid mapping. The sigmoid function $\sigma(\cdot)$ maps the adaptive persistent-consistency estimate monotonically to a token-level weight, assigning greater distillation strength to positions with stronger persistent teacher support. The trend factor $\eta_{k,i}$ further attenuates weights associated with decreasing local support. These terms produce a continuous weight that controls token contributions to the auxiliary distillation objective.

\setlength{\tabcolsep}{8pt}
\begin{table*}[t]
    \vspace{-8pt}
    \centering
    \renewcommand{\arraystretch}{0.9}
    \scalebox{0.95}
    {
    \begin{tabular*}{\textwidth}{@{\extracolsep{\fill}}lCCCCCCCCC}
    \toprule
    & \multicolumn{2}{c}{\textbf{WebShop}} & \multicolumn{7}{c}{\textbf{ALFWorld}} \\
    \cmidrule(lr){2-3} \cmidrule(lr){4-10}
    \textbf{Method}
    & \textbf{Score} & \textbf{Acc}
    & \textbf{Pick} & \textbf{Look} & \textbf{Clean} & \textbf{Heat} & \textbf{Cool} & \textbf{Pick2} & \textbf{Overall} \\
    \midrule
    \rowcolor{gray!10} \multicolumn{10}{l}{\textit{Qwen2.5-3B-Instruct}} \\
    Vanilla
        & 6.7 & 0.8
        & 44.4 & 11.1 & 6.2 & 15.4 & 28.6 & 12.5 & 21.9
        \\
    Skill-Prompt*
        & 0.2 & 0.8
        & 51.7 & 66.7 & 48.4 & 0.0 & 4.3 & 10.0 & 28.9
        \\
    OPSD
        & 11.3 & 3.1
        & 48.8 & 41.7 & 16.7 & 0.0 & 15.8 & 16.7 & 28.1
        \\
    GRPO
        & 79.8 & 63.3
        & 91.2 & 62.5 & \cellcolor{secondcolor}\underline{96.2} & 61.9 & 65.0 & 47.4 & 75.0
        \\
    Skill-GRPO
        & 77.3 & 60.9
        & 88.9 & 71.4 & 58.8 & 70.6 & 40.7 & 29.2 & 60.2
        \\
    Skill-GRPO*
        & 76.3 & \cellcolor{secondcolor}\underline{66.4}
        & 94.3 & 57.1 & \cellcolor{topcolor}\textbf{100.0} & 66.7 & 73.1 & 57.1 & 80.5
        \\
    GRPO+OPSD
        & 77.8 & \cellcolor{secondcolor}\underline{66.4}
        & \cellcolor{topcolor}\textbf{100.0} & \cellcolor{topcolor}\textbf{82.4} & 85.7 & \cellcolor{secondcolor}\underline{75.0} & 70.0 & 60.0 & 81.2
        \\
    RLSD
        & \cellcolor{secondcolor}\underline{84.4} & \cellcolor{secondcolor}\underline{66.4}
        & 87.9 & \cellcolor{secondcolor}\underline{75.0} & 90.9 & \cellcolor{secondcolor}\underline{75.0} & 73.1 & 68.4 & 79.7
        \\
    SDAR
        & 83.4 & \cellcolor{topcolor}\textbf{67.2}
        & \cellcolor{secondcolor}\underline{97.1} & 62.5 & \cellcolor{topcolor}\textbf{100.0} & 61.9 & \cellcolor{secondcolor}\underline{75.0} & \cellcolor{secondcolor}\underline{84.2} & \cellcolor{secondcolor}\underline{84.4}
        \\
    \textbf{PCSD}
        & \cellcolor{topcolor}\textbf{85.0} & \cellcolor{topcolor}\textbf{67.2}
        & \cellcolor{topcolor}\textbf{100.0} & 63.6 & 82.1 & \cellcolor{topcolor}\textbf{83.3} & \cellcolor{topcolor}\textbf{94.4} & \cellcolor{topcolor}\textbf{100.0} & \cellcolor{topcolor}\textbf{90.6}
        \\
    \midrule
    \rowcolor{gray!10} \multicolumn{10}{l}{\textit{Qwen3-1.7B-Instruct}} \\
    Vanilla
        & 46.5 & 4.7
        & 25.0 & 22.2 & 3.1 & 0.0 & 21.4 & 4.2 & 12.5
        \\
    Skill-Prompt*
        & 23.0 & 2.3
        & 10.3 & 50.0 & 16.1 & 0.0 & 0.0 & 5.0 & 9.4
        \\
    OPSD
        & 47.4 & 9.3
        & 26.3 & 33.3 & 9.1 & 0.0 & 4.5 & 5.3 & 14.1
        \\
    GRPO
        & 67.3 & 38.3
        & \cellcolor{secondcolor}\underline{71.1} & 41.7 & 36.4 & \cellcolor{secondcolor}\underline{40.0} & 31.8 & 31.6 & 46.1
        \\
    Skill-GRPO
        & 73.4 & 46.1
        & 27.6 & \cellcolor{secondcolor}\underline{54.5} & 22.7 & 27.3 & 0.0 & 19.2 & 21.1
        \\
    Skill-GRPO*
        & \cellcolor{topcolor}\textbf{80.4} & 50.0
        & 31.4 & 42.9 & 51.9 & 8.3 & 11.5 & 7.1 & 28.1
        \\
    GRPO+OPSD
        & 70.7 & 38.3
        & 38.2 & 50.0 & 30.8 & 28.6 & 30.0 & 21.1 & 32.0
        \\
    RLSD
        & 74.0 & 50.8
        & 50.0 & 37.5 & \cellcolor{secondcolor}\underline{61.5} & 19.0 & \cellcolor{topcolor}\textbf{50.0} & 21.1 & 42.2
        \\
    SDAR
        & 76.8 & \cellcolor{topcolor}\textbf{58.6}
        & \cellcolor{topcolor}\textbf{73.5} & 25.0 & \cellcolor{topcolor}\textbf{76.9} & 33.3 & 40.0 & \cellcolor{secondcolor}\underline{36.8} & \cellcolor{secondcolor}\underline{53.9}
        \\
    \textbf{PCSD}
        & \cellcolor{secondcolor}\underline{78.9} & \cellcolor{topcolor}\textbf{58.6}
        & 63.3 & \cellcolor{topcolor}\textbf{68.8} & 50.0 & \cellcolor{topcolor}\textbf{69.2} & \cellcolor{secondcolor}\underline{48.1} & \cellcolor{topcolor}\textbf{63.6} & \cellcolor{topcolor}\textbf{59.4}
        \\
    \bottomrule
    \end{tabular*}
    }
    \caption{\textbf{Performance on ALFWorld and WebShop.} For ALFWorld, we report category-wise and instance-level Overall success rates (\%).  For WebShop, we report the normalized Score and task success rate(Acc, \%) on 128 validation tasks. * denotes evaluation with skills. \sethlcolor{topcolor}\hl{\textbf{Best}} and \sethlcolor{secondcolor}\hl{\mbox{\underline{second-best}}} are highlighted.
}
\label{tab:main_results}
    \vspace{-8pt}
\end{table*}

\subsection{Training Objective}
\label{sec:training_objective}

Given the token-level weights $w_{k,i}$ derived above, let
$M=\sum_{k,i}m_{k,i}$ denote the total number of valid response tokens.
The PCSD distillation objective is
\begin{equation}
\mathcal{L}_{\mathrm{PCSD}}
=
\frac{1}{M}
\sum_{k,i}
m_{k,i}w_{k,i}\delta_{k,i},
\label{eq:pcsd_objective}
\end{equation}
where $m_{k,i}$ is the response mask and $\delta_{k,i}$ is the
teacher--student sampled log-probability gap defined in
Equation~\ref{eq:token_gap}. The teacher log-probabilities and
$w_{k,i}$ are detached during optimization. Consequently,
$\mathcal{L}_{\mathrm{PCSD}}$ induces a weighted negative-log-likelihood
gradient on student-sampled tokens. We normalize by $M$, rather than by
the sum of weights, so that $w_{k,i}$ controls both the allocation and
the effective magnitude of distillation.

We combine this auxiliary objective with GRPO:
\begin{equation}
\mathcal{L}_{\mathrm{total}}
=
\mathcal{L}_{\mathrm{GRPO}}
+
\lambda_{\methodname}\mathcal{L}_{\mathrm{PCSD}},
\label{eq:total_objective}
\end{equation}
where $\mathcal{L}_{\mathrm{GRPO}}$ uses group-relative trajectory
rewards, a clipped policy-ratio objective, and KL regularization toward
a reference policy. The trajectory-level GRPO advantage is shared by
all valid response tokens within a trajectory, whereas PCSD provides
position-specific teacher supervision through $w_{k,i}$.
The complete GRPO formulation is provided in
Appendix.

\section{Experiments}
\label{sec:experiment}

\paragraph{Benchmarks.}
We evaluate our method on two widely adopted agent benchmarks: ALFWorld~\cite{shridhar2020alfworld} and WebShop~\cite{yao2022webshop}.
ALFWorld is a text-based embodied AI benchmark across six categories of household activities: Pick and Place (Pick), Look at Obj in Light (Look), Pick Clean then Place in Recep (Clean), Pick Heat then Place in Recep (Heat), Pick Cool then Place in Recep (Cool), and Pick Two Obj and Place (Pick2). 
WebShop simulates realistic e-commerce scenarios where agents navigate product catalogs to purchase items matching user specifications; we use 1,000 training tasks and evaluate on 128 fixed validation instances following~\cite{feng2026group}.  The task spans both information-seeking and decision-making, with a large action space of heterogeneous products.

\paragraph{Baselines.}
We compare PCSD with prompting, RL, and self-distillation baselines on two base models. Vanilla uses no skills, while Skill-Prompt and Skill-GRPO use skills retrieved by keyword matching during inference and training, respectively. GRPO~\cite{shao2024deepseekmath} uses group-relative rewards, and OPSD~\cite{zhao2026self} uses self-distillation alone. GRPO+OPSD, RLSD~\cite{yang2026self}, SDAR~\cite{lu2026self}, and PCSD combine GRPO with token-level privileged-teacher distillation. These four methods share the student, teacher-side skills, training budget, and evaluation protocol, differing only in their weighting rules. Shared hyperparameters are provided in the appendix, and * denotes inference-time skill use. We reimplement SDAR following its original method and report results under this shared setup.

\paragraph{Implementation Details.}
We conduct experiments using Qwen2.5-3B-Instruct and Qwen3-1.7B-Instruct backbones on 8$\times$A100 GPUs.
For ALFWorld, we adopt the GiGPO~\citep{feng2026group} data split, use a batch size of 128 (16 tasks × 8 rollouts per prompt), and set the maximum prompt length to 2,048 tokens. For WebShop, each batch contains 16 tasks with eight rollouts per prompt, and the maximum prompt length is set to 4,096 tokens. All experiments run for 150 steps with AdamW with gradient clipping threshold 1.0. For \methodname hyperparameters, we set the distillation coefficient $\lambda_{\methodname}=0.01$, sigmoid sharpness parameter $\beta_{\mathrm{gate}}=5.0$. Other settings are provided in the Appendix.

\paragraph{Evaluation metrics.} For ALFWorld, we report category-wise and overall success rates, with the latter computed across all evaluation instances. For WebShop, we follow the official protocol and report the normalized score, measuring product–specification alignment, and the task success rate.

\subsection{Main Results}
\paragraph{Overall Performance.}
Table~\ref{tab:main_results} reports the main results. On ALFWorld, \methodname achieves the highest Overall success rates of 90.6\% and 59.4\% with Qwen2.5-3B-Instruct and Qwen3-1.7B-Instruct, respectively. These results exceed GRPO by 15.6 and 13.3 percentage points and the strongest competing distillation baseline, SDAR, by 6.2 and 5.5 percentage points. The category-wise results show particularly strong performance on Heat and Pick2 for both backbones. On WebShop, \methodname achieves the highest Score of 85.0 with Qwen2.5-3B-Instruct and ties with SDAR for the highest Acc on both backbones. With Qwen3-1.7B-Instruct, its Score of 78.9 ranks second behind Skill-GRPO*. Overall, the results support the effectiveness of \methodname's adaptive weighting mechanism, which yields consistent gains across both evaluated model scales.

\paragraph{Performance without Inference-Time Skill Retrieval.}
The effect of inference-time skill retrieval varies across backbones and tasks. Skill-Prompt improves over Vanilla on ALFWorld with Qwen2.5-3B-Instruct, but performs worse with Qwen3-1.7B-Instruct and on WebShop. Similarly, Skill-GRPO* benefits from retrieved skills at evaluation time but remains below \methodname on the ALFWorld Overall metric for both backbones. In contrast, \methodname requires no external skill retrieval during evaluation, indicating that its reported performance is achieved by the trained student policy alone.

\paragraph{Comparison with Hybrid Baselines.}
Distillation and hybrid baselines show mixed results.
OPSD performs poorly on ALFWorld, while GRPO+OPSD improves on Qwen2.5-3B-Instruct but degrades on Qwen3-1.7B-Instruct, indicating that naive RL-distillation combination can introduce optimization interference.
Although RLSD and SDAR are stronger baselines, PCSD still achieves the best aggregate ALFWorld results on both models.
Fine-grained subtask results further show that PCSD performs robustly across diverse ALFWorld tasks, especially on Heat, Cool, and Pick2.

\subsection{Training Dynamics}
\label{sec:training_dynamics}

\begin{figure}[t]

    \centering
    \includegraphics[width=\linewidth]{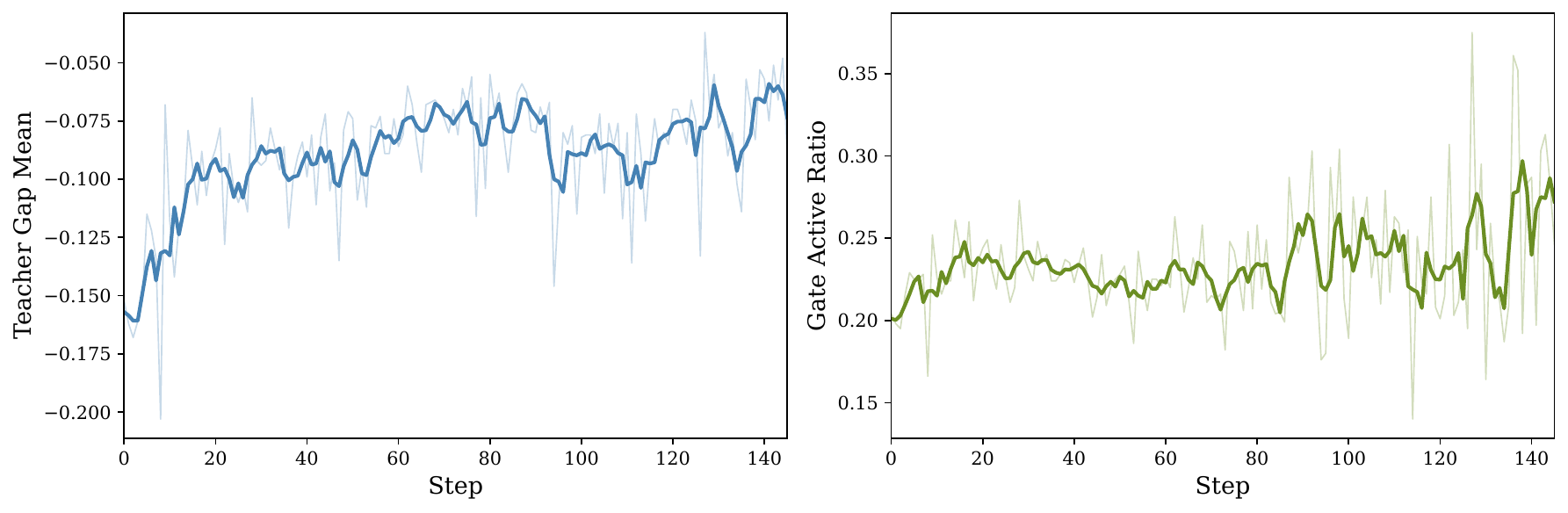}
    \caption{Training dynamics.
    Average teacher–student gap (left) and gate activation ratio (right) for Qwen2.5-3B-Instruct on ALFWorld. Translucent and solid curves show raw and smoothed values, respectively.
    }
    \label{fig:training_dynamics}
    \vspace{-8pt}
\end{figure}

To examine the adaptive behavior of \methodname, we track the
teacher--student log-probability gap and gate activation ratio during RL
training on ALFWorld (Figure~\ref{fig:training_dynamics}). The mean gap
remains negative but gradually increases, indicating that student-generated tokens receive progressively stronger relative support from the teacher. Meanwhile, 20\%--28\% of valid tokens have $w_{k,i} > 0.5$, indicating concentrated rather than hard-selective distillation. Together, these dynamics demonstrate that \methodname adaptively reallocates token-level supervision as student policy evolves.

\subsection{Generalization to Unseen Environments}
\label{sec:unseen_generalization}

\begin{figure}[t]

    \centering
    \includegraphics[width=0.9\linewidth]{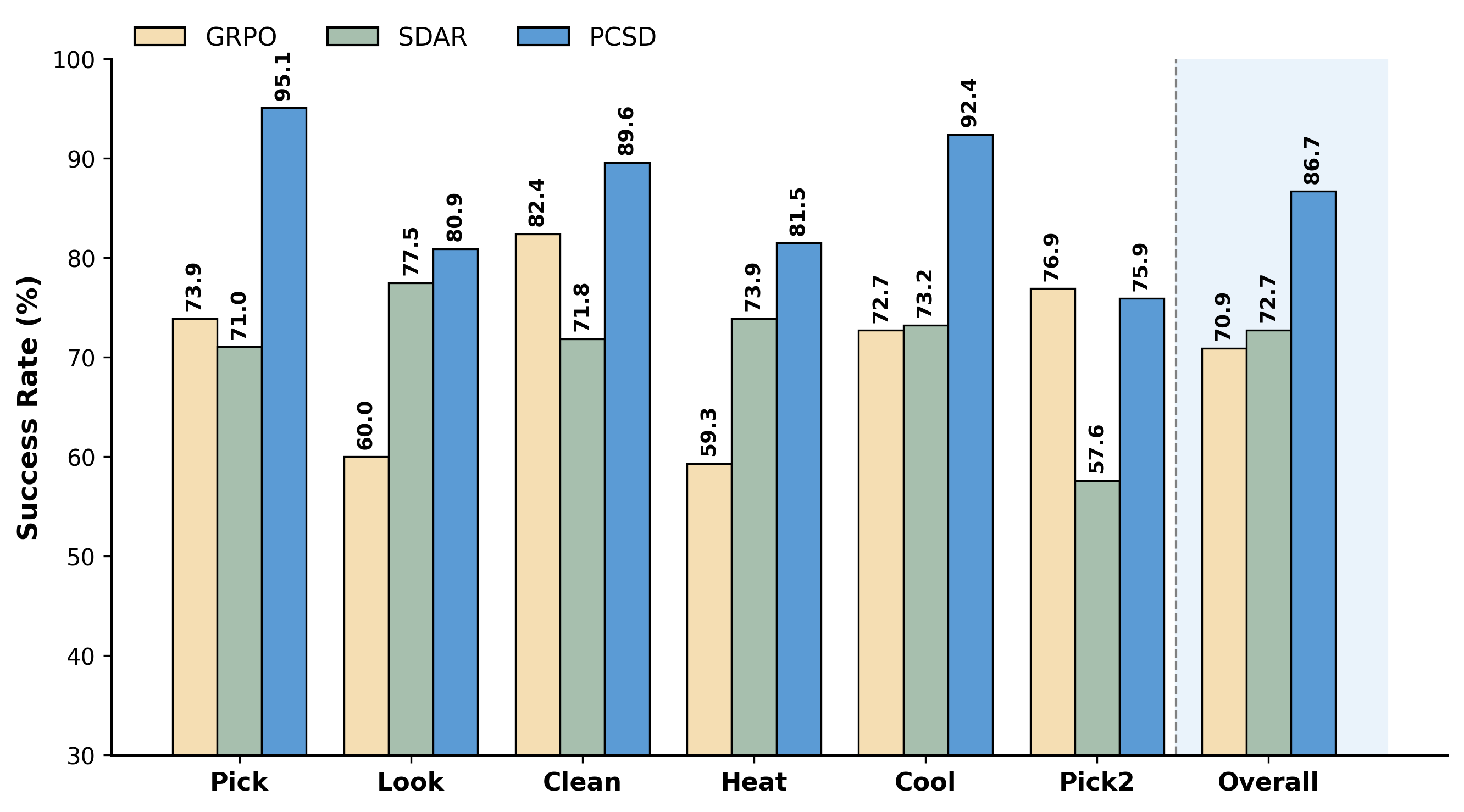}
    \caption{Generalization to unseen ALFWorld environments.
    Category-wise and Overall success rates (\%) of SDAR, GRPO and \methodname on the ALFWorld unseen split.}
    \label{fig:unseen_generalization}
    \vspace{-8pt}
\end{figure}

As shown in Figure~\ref{fig:unseen_generalization}, \methodname consistently outperforms both GRPO and SDAR across the ALFWorld unseen-split categories, achieving the highest success rate on nearly every subtask. Its Overall success rate reaches 86.7\%, well above GRPO's 70.9\% and SDAR's 72.7\%.  These results suggest that the benefits of persistent-consistency weighting extend to unseen environment configurations without requiring privileged skills during evaluation.

\begin{table}[h]
\centering
\footnotesize
\setlength{\tabcolsep}{2.0pt}
\renewcommand{\arraystretch}{1.10}
\begin{tabular}{@{}lccccccc@{}}
\toprule
\textbf{Method}
& \textbf{Pick}
& \textbf{Look}
& \textbf{Clean}
& \textbf{Heat}
& \textbf{Cool}
& \textbf{Pick2}
& \textbf{Overall} \\
\midrule

\rowcolor{gray!10} PCSD
& 100.0 & 63.6 & 82.1 & 83.3 & 94.4 & 100.0 & \textbf{90.6} \\
\midrule
Fixed $N=1$
& 90.5 & 82.4 & 100.0 & 69.2 & 84.0 & 64.0 & 82.8 \\
Fixed $N=4$
& 100.0 & 63.6 & 100.0 & 83.3 & 94.4 & 65.1 & 88.3 \\
w/o trend
& 96.4 & 85.7 & 90.9 & 91.7 & 72.7 & 64.0 & 83.6 \\
w/o decay
& 86.1 & 63.6 & 96.4 & 83.3 & 100.0 & 69.6 & 85.1 \\
\bottomrule
\end{tabular}
\caption{\textbf{Component ablation on ALFWorld.}
Success rates (\%) using Qwen2.5-3B-Instruct. $N=1$ denotes pointwise
weighting, $N=4$ a fixed local window, and Overall the success rate across all evaluation instances.}
\label{tab:component_ablation}
\vspace{-10pt}
\end{table}

\subsection{Ablation Studies}

\paragraph{Component Ablation.}
Table~\ref{tab:component_ablation} evaluates the key components of \methodname. Replacing adaptive aggregation with fixed windows, removing trend modulation, or substituting exponential decay with uniform weighting consistently reduces Overall performance. Although some variants perform better on individual categories, the complete \methodname achieves the best Overall result, validating the complementary contributions of adaptive aggregation, trend modulation, and proximity-aware weighting.

\paragraph{Sensitivity to the Distillation Coefficient.}
Table~\ref{tab:lambda_ablation} evaluates the effect of $\lambda_{\methodname}$. Removing the distillation objective yields an Overall success rate of 75.0\%. Setting $\lambda_{\methodname}$ to 0.005 and 0.01 improves the result to 87.5\% and 90.6\%, respectively, supporting the benefit of combining token-level teacher supervision with trajectory-level rewards. Increasing $\lambda_{\methodname}$ to 0.05 reduces the Overall success rate to 83.6\%, indicating that excessive distillation strength can weaken the balance between teacher supervision and reward optimization. Performance is non-monotonic over the evaluated values, with $\lambda_{\methodname}=0.01$ achieving the best Overall result.

\begin{table}[h]
\centering

\small
\setlength{\tabcolsep}{3pt}
\renewcommand{\arraystretch}{1.08}
\begin{tabular}{@{}lccccccc@{}}
\toprule
& \multicolumn{7}{c}{\textbf{ALFWorld}} \\
\cmidrule(lr){2-8}
$\lambda_{\methodname}$
& \textbf{Pick} & \textbf{Look} & \textbf{Clean} & \textbf{Heat}
& \textbf{Cool} & \textbf{Pick2} & \textbf{Overall} \\
\midrule

\rowcolor{gray!10} 0.01
& 100.0 & 63.6 & 82.1 & 83.3 & 94.4 & 100.0 & \textbf{90.6} \\
\midrule
0.005
& 96.0 & 90.0 & 86.4 & 79.1 & 71.4 & 96.0 & 87.5 \\
0.05
& 100.0 & 72.7 & 83.9 & 100.0 & 72.0 & 71.4 & 83.6 \\
0.0
& 91.2 & 62.5 & 96.2 & 61.9 & 65.0 & 47.4 & 75.0 \\
\bottomrule
\end{tabular}

\caption{
\textbf{Sensitivity to the distillation coefficient.}
We report category-wise and overall success rates (\%) on ALFWorld using
Qwen2.5-3B-Instruct.
}
\label{tab:lambda_ablation}
\vspace{-10pt}
\end{table}

\section{Discussion}

PCSD uses fixed hyperparameters for local aggregation and gating and keeps the privileged teacher frozen. These choices isolate the effect of persistent-consistency weighting and avoid coupling policy optimization with a changing teacher or weighting mechanism. The trade-off is reduced adaptation to evolving trajectory statistics and variations in teacher reliability. Future work could learn context-dependent aggregation and gating parameters from trajectory statistics, teacher uncertainty, and environmental feedback. Another direction is self-evolving distillation, in which the student, teacher, credibility estimator, and skill repository co-evolve through online interaction.

\section{Conclusion}

We propose \methodname, an on-policy self-distillation framework that weights token-level supervision based on persistent local support from the teacher. By integrating adaptive exponential-decay aggregation, one-sided trend modulation, and continuous gating, \methodname incorporates evidence across multiple tokens while preserving positional specificity. Experiments demonstrate consistent improvements over outcome-only RL and existing self-distillation baselines.

\bibliography{aaai2027}

\clearpage

\appendix
\section{Appendix}

This appendix provides additional details on the formulation, implementation,
and analysis of \methodname{}. We first describe the construction of the
privileged teacher, skill retrieval, information isolation, and the prompt
templates used in both environments. We then present the complete GRPO
objective and PCSD training procedure, followed by analyses of the local
bias--variance trade-off, exponential aggregation, effective window length,
and one-sided trend modulation. Finally, we report the training settings,
benchmark protocols, and diagnostic results on weight robustness and
teacher-quality perturbations. These materials complement the main paper by
clarifying the implementation and supporting the reproducibility and
interpretation of our results.


\subsection{Teacher and Privileged Skills}
\label{app:teacher_skills}

\paragraph{Teacher construction.} We adopt an asymmetric teacher--student training setup. The privileged teacher $\pi_T$ is initialized as a separate frozen copy of the same base checkpoint used to initialize the student. It remains fixed throughout reinforcement learning: no gradient is propagated through its parameters, and it is not included in the optimizer. For every trajectory sampled by the student, the teacher evaluates the same student-generated tokens through teacher forcing and provides token-level conditional probabilities for the PCSD objective. Consequently, all optimization gradients act only on the student policy. The privileged teacher is also distinct from the frozen reference policy used for GRPO regularization: the former provides skill-conditioned distillation signals, whereas the latter only constrains policy deviation.

\paragraph{Privileged skills and retrieval.} The skill repository contains general interaction rules and task-type-specific procedural knowledge. These skills describe reusable action strategies, such as object manipulation or navigation procedures, but do not contain solutions to individual evaluation instances, expert trajectories, target-object locations, future observations, or hidden environment states. For each trajectory, we retrieve task-type skills by keyword matching against the observable task instruction and combine them with the general skills. If no task-type skill is matched, only the general skills are provided. Thus, the privileged context supplies task-level procedural priors rather than instance-level answers. The repository is constructed without access to validation trajectories. ALFWorld uses its designated training, seen-validation, and unseen-validation splits, while WebShop uses 1,000 training tasks and 128 fixed validation instances.

\paragraph{Prompt construction and information isolation.} The student conditions only on the original task instruction, observable environment feedback, and its causal interaction history; privileged skills are never included in the student input during either training or evaluation. The teacher receives the same observable interaction prefix together with the retrieved skills and scores the exact tokens generated by the student. For token $y_{k,i}$, the teacher and student input contexts therefore differ only in the additional skill information available to the teacher, while the evaluated token and causal interaction history remain aligned. No alternative action is sampled from the teacher, avoiding comparison bias caused by different generations. Skill retrieval depends only on the initial observable task description and does not use rewards, future observations, hidden states, or the student's free-form reasoning. These constraints prevent instance-specific privileged or future information from entering the teacher context beyond the observable context shared with the student. Thus, evaluation performance reflects knowledge internalized by the student rather than continued reliance on external skills.

\subsection{Prompt Templates}
\label{app:prompt_templates}

\paragraph{ALFWorld.}
For ALFWorld, the agent is prompted as an embodied decision-maker operating
in the ALFRED environment. At each interaction step, it receives the task
objective, a bounded history of recent observations and executed actions, the
current textual environment observation, and the set of currently admissible
actions. The prompt requires the agent to first produce step-by-step reasoning
enclosed within
\texttt{\textless think\textgreater...\textless/think\textgreater} tags,
followed by exactly one executable action enclosed within
\texttt{\textless action\textgreater...\textless/action\textgreater} tags.
The action must be selected from the admissible action set, which includes
navigation, object acquisition, container manipulation, object-state
transformation, and placement operations.

At the initial step, the interaction history is omitted, while the task
description, current observation, and admissible actions are retained. At
subsequent steps, the prompt additionally includes the current interaction
index and the most recent $H$ observation--action pairs, where $H$ is the
configured history length. The prompt template is:

\begin{quote}
\small
\ttfamily
\raggedright
You are an expert agent operating in the ALFRED Embodied Environment. Your
task is to: \{task\_description\}

\par\medskip
Prior to this step, you have already taken \{step\_count\} step(s). Below are
the most recent \{history\_length\} observations and the corresponding actions
you took: \{action\_history\}

\par\medskip
You are now at step \{current\_step\} and your current observation is:
\{current\_observation\}

\par\medskip
Your admissible actions of the current situation are:
[\{admissible\_actions\}].

\par\medskip
Now it's your turn to take an action.

\par
You should first reason step-by-step about the current situation. This
reasoning process MUST be enclosed within
\textless think\textgreater\ \textless/think\textgreater\ tags.

\par
Once you've finished your reasoning, you should choose an admissible action
for current step and present it within
\textless action\textgreater\ \textless/action\textgreater\ tags.
\end{quote}

This prompt exposes the causal structure required by long-horizon embodied
tasks. For example, the agent may need to locate an object, acquire it, apply
the required transformation, such as heating, cooling, or cleaning, navigate
to the destination receptacle, and place the transformed object.

\paragraph{WebShop.}
For WebShop, the agent is prompted as an autonomous e-commerce agent. At each
step, it receives the shopping request, including the product category and
attribute constraints, a bounded history of recent page observations and
actions, the current text-rendered web page, and the set of currently
available search and click operations. The prompt asks the agent to reason
about which admissible action best advances the shopping objective and then
output exactly one action in the required format.

At the initial step, the interaction history is omitted. At subsequent steps,
the prompt additionally includes the current step index and the most recent
$H$ observation--action pairs. The prompt template is:

\begin{quote}
\small
\ttfamily
\raggedright
You are an expert autonomous agent operating in the WebShop e-commerce
environment.

\par
Your task is to: \{task\_description\}.

\par\medskip
Prior to this step, you have already taken \{step\_count\} step(s). Below are
the most recent \{history\_length\} observations and the corresponding actions
you took: \{action\_history\}

\par\medskip
You are now at step \{current\_step\} and your current observation is:
\{current\_observation\}.

\par\medskip
Your admissible actions of the current situation are:

\par
[\{available\_actions\}].

\par\medskip
Now it's your turn to take one action for the current step.

\par
You should first reason step-by-step about the current situation, then think
carefully which admissible action best advances the shopping goal. This
reasoning process MUST be enclosed within
\textless think\textgreater\ \textless/think\textgreater\ tags.

\par
Once you've finished your reasoning, you should choose an admissible action
for current step and present it within
\textless action\textgreater\ \textless/action\textgreater\ tags.
\end{quote}

The WebShop action space is normalized into two forms:
\texttt{search[\textless your query\textgreater]} for issuing a product query
and \texttt{click[\textless item\textgreater]} for interacting with products,
filters, variants, navigation controls, or the purchase button. This
formulation requires the agent to jointly perform product retrieval,
attribute verification, variant selection, and final purchase. If the
assembled prompt exceeds the configured length threshold, the interaction
history is removed and the corresponding no-history prompt is used.

\paragraph{Student and Teacher Inputs.}
During \methodname{} training, the student receives the standard environment
prompt described above. The teacher receives the same task description,
interaction history, current observation, and admissible actions, with
task-relevant privileged skill information prepended to the prompt. The
teacher evaluates the same response tokens generated by the student rather
than sampling a separate action sequence. Thus, the teacher--student
probability comparison uses identical environment context and
student-generated tokens, with the retrieved skill information provided only
to the teacher.

\subsection{GRPO Objective}
\label{app:grpo_objective}

We use the following GRPO objective in all reinforcement-learning
experiments. For each input, the rollout policy
$\pi_{\theta_{\mathrm{old}}}$ samples a group of $G$ complete interaction
trajectories. We use $g\in\{1,\ldots,G\}$ to index trajectories and $i$
to index student-generated response tokens. Let $m_{g,i}\in\{0,1\}$
denote the response-token mask and $M_g=\sum_i m_{g,i}$.

Given the trajectory rewards $\{R_g\}_{g=1}^{G}$, we compute
\begin{equation}
\mu_R=\frac{1}{G}\sum_{g=1}^{G}R_g,
\qquad
\sigma_R=
\sqrt{\frac{1}{G}\sum_{g=1}^{G}(R_g-\mu_R)^2},
\end{equation}
and define the group-relative advantage as
\begin{equation}
A_g=
\frac{R_g-\mu_R}
     {\sigma_R+\epsilon_{\mathrm{num}}},
\label{eq:grpo_advantage}
\end{equation}
where $\epsilon_{\mathrm{num}}>0$ ensures numerical stability. The same
trajectory-level advantage is assigned to all valid response tokens in
trajectory $g$.

Let $h_{g,i}^{S}$ denote the student-visible context preceding token
$y_{g,i}$. For numerical stability, the importance ratio is computed as
\begin{equation}
\rho_{g,i}
=
\exp\left(
\log\pi_\theta(y_{g,i}\mid h_{g,i}^{S})
-
\log\pi_{\theta_{\mathrm{old}}}
(y_{g,i}\mid h_{g,i}^{S})
\right),
\label{eq:grpo_ratio}
\end{equation}
with
\begin{equation}
\bar{\rho}_{g,i}
=
\operatorname{clip}
\left(
\rho_{g,i},
1-\epsilon_{\mathrm{clip}},
1+\epsilon_{\mathrm{clip}}
\right).
\label{eq:grpo_clipped_ratio}
\end{equation}

We estimate the token-level KL penalty using
\begin{equation}
\begin{aligned}
\Delta_{g,i}
&=
\log
\frac{\pi_{\mathrm{ref}}(y_{g,i}\mid h_{g,i}^{S})}
     {\pi_{\theta}(y_{g,i}\mid h_{g,i}^{S})},\\
\widehat D_{\mathrm{KL}}^{\,g,i}
&=
\exp(\Delta_{g,i})-\Delta_{g,i}-1.
\end{aligned}
\label{eq:grpo_kl}
\end{equation}

The resulting GRPO loss is
\begin{align}
\mathcal{L}_{\mathrm{GRPO}}
&=
-\frac{1}{G}
\sum_{g=1}^{G}
\frac{1}{M_g}
\sum_i m_{g,i}
\min\left(
\rho_{g,i}A_g,\,
\bar{\rho}_{g,i}A_g
\right)
\notag\\
&\quad+
\frac{\beta_{\mathrm{KL}}}{G}
\sum_{g=1}^{G}
\frac{1}{M_g}
\sum_i m_{g,i}
\widehat D_{\mathrm{KL}}^{\,g,i}.
\label{eq:grpo_objective}
\end{align}

Here, $\epsilon_{\mathrm{clip}}$ is the policy-ratio clipping coefficient
and $\beta_{\mathrm{KL}}$ controls regularization toward the frozen
reference policy. The response mask excludes prompt, environment, and
padding tokens. The rollout policy $\pi_{\theta_{\mathrm{old}}}$ is
updated between rollout iterations, whereas $\pi_{\mathrm{ref}}$ remains
fixed throughout training and is distinct from the privileged teacher.

\subsection{Training Procedure}
\label{app:training_procedure}

Algorithm~\ref{alg:pcsd_training} summarizes the complete training
procedure. At each update, the current student is copied to the rollout
policy, which collects $G$ trajectories for every sampled task without
access to privileged skills. The resulting trajectory rewards are
normalized within each task group to obtain GRPO advantages. The frozen
teacher then evaluates the same student-generated tokens under a
skill-augmented context. No response is resampled from the teacher.

PCSD weights are computed independently within each student response;
local windows never cross interaction-turn boundaries. The
teacher--student gaps used for variance estimation, aggregation, trend
analysis, and gating are detached from the computation graph. However,
the student log-probabilities in the PCSD loss remain differentiable, so
the loss produces a weighted negative-log-likelihood gradient. The
GRPO loss already includes regularization toward the frozen reference
policy. Only the student parameters are updated, while the teacher,
reference policy, skill retriever, and PCSD weights remain fixed during
each optimization step.

\begin{algorithm*}[t]
\caption{\methodname{} Training}
\label{alg:pcsd_training}
{
\small
\begin{algorithmic}[1]
\REQUIRE Student $\pi_\theta$, frozen teacher $\pi_T$, frozen reference
policy $\pi_{\mathrm{ref}}$, task set $\mathcal D$, skill retriever
$\mathcal R$
\REQUIRE Group size $G$, update steps $K$, and PCSD hyperparameters
\ENSURE Trained student policy $\pi_\theta$

\FOR{$u=1,\ldots,K$}
    \STATE Sample a task batch $\{x_b\}_{b=1}^{B}$ from $\mathcal D$
    \STATE Set $\theta_{\mathrm{old}}\leftarrow\theta$

    \FOR{each task $x_b$}
        \STATE Retrieve task-level skills
        $\mathcal S_b\leftarrow\mathcal R(x_b)$
        \STATE Sample $G$ trajectories
        $\tau_{b,g}\sim\pi_{\theta_{\mathrm{old}}}(\cdot\mid x_b)$
        without skills
        \STATE Compute returns $R_{b,g}$ and group-normalized advantages
        $\widehat A_{b,g}$

        \FOR{each trajectory $\tau_{b,g}$ and response turn $t$}
            \STATE Let $y_{b,g,t}$ be the student-generated response
            with mask $m_{b,g,t,i}$

            \FOR{each valid token $y_{b,g,t,i}$}
                \STATE Compute student log-probability
                $\ell^S_{b,g,t,i}
                =\log\pi_\theta(y_{b,g,t,i}\mid h^S_{b,g,t,i})$
                \STATE Compute teacher log-probability
                $\ell^T_{b,g,t,i}
                =\log\pi_T(y_{b,g,t,i}\mid h^T_{b,g,t,i},\mathcal S_b)$
                \STATE Set the detached gap
                $\delta_{b,g,t,i}
                =\operatorname{sg}(\ell^T_{b,g,t,i}-\ell^S_{b,g,t,i})$
            \ENDFOR

            \FOR{each valid token position $i$}
                \STATE Compute local variance $V_{b,g,t,i}$ over the
                masked $N_{\max}$ window
                \STATE $r_{b,g,t,i}\leftarrow
                \operatorname{clip}\!\left(
                \frac{V_{b,g,t,i}-\tau_{\mathrm{low}}}
                {\tau_{\mathrm{high}}-\tau_{\mathrm{low}}},
                0,1\right)$
                \STATE Compute exponentially weighted estimates
                $\bar\delta^{(N_{\min})}_{b,g,t,i}$ and
                $\bar\delta^{(N_{\max})}_{b,g,t,i}$
                \STATE $\bar\delta^{\mathrm{adaptive}}_{b,g,t,i}
                \leftarrow
                (1-r_{b,g,t,i})\bar\delta^{(N_{\min})}_{b,g,t,i}
                +r_{b,g,t,i}\bar\delta^{(N_{\max})}_{b,g,t,i}$
                \STATE Compute the mask-aware OLS slope
                $s_{b,g,t,i}$ over the $N_{\max}$ window; set it to
                zero when fewer than two valid tokens are available
                \STATE Compute the response-level scale
                $\delta_{\mathrm{scale},b,g,t}$
                \STATE $\eta_{b,g,t,i}\leftarrow
                \operatorname{clip}\!\left(
                1-\gamma\operatorname{ReLU}
                \left(-s_{b,g,t,i}/\delta_{\mathrm{scale},b,g,t}\right),
                0,1\right)$
                \STATE $w_{b,g,t,i}\leftarrow
                \operatorname{sg}\!\left[
                \sigma\!\left(
                \beta_{\mathrm{gate}}
                \bar\delta^{\mathrm{adaptive}}_{b,g,t,i}
                \right)
                \eta_{b,g,t,i}\right]$
            \ENDFOR
        \ENDFOR
    \ENDFOR

    \STATE $M\leftarrow\sum_{b,g,t,i}m_{b,g,t,i}$
    \STATE $\displaystyle
    \mathcal L_{\mathrm{PCSD}}\leftarrow
    \frac{1}{M}\sum_{b,g,t,i}
    m_{b,g,t,i}w_{b,g,t,i}
    \left[\operatorname{sg}(\ell^T_{b,g,t,i})
    -\ell^S_{b,g,t,i}\right]$
    \STATE Compute $\mathcal L_{\mathrm{GRPO}}$ using
    $\widehat A_{b,g}$, $\pi_{\theta_{\mathrm{old}}}$, and
    $\pi_{\mathrm{ref}}$
    \STATE $\mathcal L_{\mathrm{total}}\leftarrow
    \mathcal L_{\mathrm{GRPO}}
    +\lambda_{\mathrm{PCSD}}\mathcal L_{\mathrm{PCSD}}$
    \STATE Update only $\theta$ using
    $\nabla_\theta\mathcal L_{\mathrm{total}}$
\ENDFOR

\RETURN $\pi_\theta$
\end{algorithmic}
}
\end{algorithm*}

\subsection{Bias--Variance Trade-off in High-Variance Regions}
\label{app:variance_tradeoff}

The variance-conditioned aggregation in \methodname{} reflects a local
bias--variance trade-off rather than treating high variance as direct evidence
of teacher unreliability. A short window preserves abrupt token-level changes
but is sensitive to isolated fluctuations, whereas a longer window provides a
more stable estimate at the risk of smoothing meaningful transitions.

Let $\delta_t$ denote the detached teacher--student log-probability gap at
token $t$. For a window of size $N$, the mask-aware estimate is
\begin{equation}
\bar{\delta}_t^{(N)}
=
\sum_{i=0}^{N-1}a_{t,i}^{(N)}\delta_{t+i},
\qquad
a_{t,i}^{(N)}
=
\frac{\alpha^i m_{t+i}}
{\sum_{j=0}^{N-1}\alpha^j m_{t+j}},
\label{eq:tradeoff_estimator}
\end{equation}
where $m_{t+i}$ is the valid-token mask. Hence, aggregation does not cross
response boundaries.

To characterize this estimator, suppose
$\delta_{t+i}=\mu_{t+i}+\varepsilon_{t+i}$, where
$\mathrm{E}[\varepsilon_{t+i}]=0$,
$\operatorname{Var}[\varepsilon_{t+i}]\leq\sigma^2$, and the local signal
satisfies $|\mu_{t+i}-\mu_t|\leq Li$. Define
\begin{equation}
d_t^{(N)}
=
\sum_{i=0}^{N-1}i\,a_{t,i}^{(N)},
\qquad
N_{\mathrm{eff},t}^{(N)}
=
\frac{1}
{\sum_{i=0}^{N-1}\left(a_{t,i}^{(N)}\right)^2}.
\label{eq:tradeoff_statistics}
\end{equation}
Under independent local noise, the mean-squared error is bounded by
\begin{equation}
\mathrm{E}
\left[
\left(
\bar{\delta}_t^{(N)}-\mu_t
\right)^2
\right]
\leq
L^2\left(d_t^{(N)}\right)^2
+
\frac{\sigma^2}{N_{\mathrm{eff},t}^{(N)}}.
\label{eq:tradeoff_mse}
\end{equation}
The first term represents temporal smoothing bias, whereas the second
represents sensitivity to local noise. A longer window generally increases
the effective sample size and reduces the second term, but may increase the
first when teacher support changes rapidly. With $N_{\max}=8$ and
$\alpha=0.8$, the effective sample size is approximately $6.41$, while the
average temporal displacement is only about $2.39$ tokens. Exponential decay
therefore improves stability while retaining a preference for nearby tokens.

\methodname{} balances the two estimates through
\begin{equation}
\begin{aligned}
r_t
&=
\operatorname{clip}
\left(
\frac{V_t-\tau_{\mathrm{low}}}
{\tau_{\mathrm{high}}-\tau_{\mathrm{low}}},
0,1
\right),\\
\bar{\delta}_t^{\mathrm{ad}}
&=
(1-r_t)\bar{\delta}_t^{(N_{\min})}
+
r_t\bar{\delta}_t^{(N_{\max})}.
\end{aligned}
\label{eq:tradeoff_adaptive}
\end{equation}
Continuous interpolation avoids abrupt changes around the variance thresholds.
Low-variance regions retain more token-level detail, while high-variance
regions place greater weight on the more stable long-window estimate.

Importantly, $V_t$ may reflect both stochastic fluctuations and genuine
changes in teacher support. Therefore, high variance does not always favor
additional smoothing. \methodname{} limits this risk using a small maximum
window, exponential decay, and continuous interpolation. Nevertheless, abrupt
changes within a short span may still be partially smoothed.

The sigmoid gate also transfers estimation stability to the token weights.
Since $|\sigma'(x)|\leq1/4$, its sigmoid component satisfies
\begin{equation}
\left|
\sigma\left(\beta_{\mathrm{gate}}\bar{\delta}_t^{\mathrm{ad}}\right)
-
\sigma\left(\beta_{\mathrm{gate}}\mu_t\right)
\right|
\leq
\frac{\beta_{\mathrm{gate}}}{4}
\left|
\bar{\delta}_t^{\mathrm{ad}}-\mu_t
\right|.
\label{eq:tradeoff_gate_stability}
\end{equation}
Thus, a more stable gap estimate produces a more stable distillation weight,
although a larger $\beta_{\mathrm{gate}}$ also increases sensitivity to
estimation error.

This analysis characterizes the local estimator rather than establishing a
global convergence guarantee. Any residual smoothing affects only the
auxiliary distillation weights; the trajectory-level GRPO objective remains
unchanged. Future work may learn context-dependent window sizes, decay
factors, and variance thresholds, allowing the trade-off to adapt across
different reasoning stages and training dynamics.

\subsection{Further Analysis of Local Aggregation}
\label{app:local_aggregation_analysis}

\paragraph{Denoising effect of exponential aggregation.}
Consider a complete forward window of length $N$ with normalized exponential
weights
\begin{equation}
a_i^{(N)}
=
\frac{\alpha^i}
{\sum_{j=0}^{N-1}\alpha^j},
\qquad
\bar{\delta}_t^{(N)}
=
\sum_{i=0}^{N-1}
a_i^{(N)}\delta_{t+i}.
\label{eq:exp_aggregation_analysis}
\end{equation}
Suppose that the observed gap is
$\delta_{t+i}=\mu_t+\varepsilon_{t+i}$ within a locally stationary region,
where $\mathrm{E}[\varepsilon_{t+i}]=0$ and
$\operatorname{Var}[\varepsilon_{t+i}]=\sigma^2$. If the noise terms are
independent, then
\begin{equation}
\operatorname{Var}
\left[
\bar{\delta}_t^{(N)}
\right]
=
\sigma^2
\sum_{i=0}^{N-1}
\left(a_i^{(N)}\right)^2
=
\frac{\sigma^2}
{N_{\mathrm{eff}}(N,\alpha)}.
\label{eq:exp_variance_reduction}
\end{equation}
Thus, exponential aggregation reduces the variance of an individual
token-level gap while preserving temporal locality.

The same result also explains the response to an isolated perturbation.
If the gap at offset $j$ is changed by $\xi$, the aggregated signal changes by
\begin{equation}
\left|
\bar{\delta}_t^{(N)\prime}
-
\bar{\delta}_t^{(N)}
\right|
=
a_j^{(N)}|\xi|
=
\frac{(1-\alpha)\alpha^j}
{1-\alpha^N}
|\xi|.
\label{eq:isolated_gap_influence}
\end{equation}
The influence of an isolated fluctuation is therefore bounded and decreases exponentially with its distance from the current token. This reduces the influence of a distant isolated gap on the local credit signal.

Token-level noise may be correlated in practice. Under an equicorrelation
model with pairwise correlation coefficient $c$, the variance becomes
\begin{equation}
\operatorname{Var}
\left[
\bar{\delta}_t^{(N)}
\right]
=
\sigma^2
\left[
c+
\frac{1-c}
{N_{\mathrm{eff}}(N,\alpha)}
\right].
\label{eq:correlated_noise_variance}
\end{equation}
Positive correlation reduces the attainable variance reduction, but the
aggregation still suppresses the uncorrelated component of local noise.

\paragraph{Effective window length.}
The nominal window size $N$ does not fully describe the amount of averaging,
because exponential weights contribute unequally. A more informative measure
is
\begin{align}
N_{\mathrm{eff}}(N,\alpha)
&=
\frac{
\left(
\sum_{i=0}^{N-1}\alpha^i
\right)^2
}{
\sum_{i=0}^{N-1}\alpha^{2i}
}
\notag\\
&=
\frac{
(1+\alpha)(1-\alpha^N)
}{
(1-\alpha)(1+\alpha^N)
}.
\label{eq:effective_window_length}
\end{align}
It satisfies $1\leq N_{\mathrm{eff}}\leq N$. When $\alpha$ approaches zero,
the estimate is dominated by the current token and
$N_{\mathrm{eff}}\rightarrow1$. When $\alpha$ approaches one, the weights
become uniform and $N_{\mathrm{eff}}\rightarrow N$.

For the setting used in our experiments,
$N_{\max}=8$ and $\alpha=0.8$, giving
\begin{equation}
N_{\mathrm{eff}}(8,0.8)
\approx 6.41.
\label{eq:effective_window_value}
\end{equation}
The corresponding weighted temporal displacement is
\begin{equation}
d_N
=
\frac{
\sum_{i=0}^{N-1}i\alpha^i
}{
\sum_{i=0}^{N-1}\alpha^i
},
\qquad
d_8\approx2.39.
\label{eq:effective_temporal_displacement}
\end{equation}
Although the nominal window contains eight tokens, its center of influence is
only about $2.39$ tokens ahead of the current position. The window therefore
provides substantial averaging without treating all eight positions
uniformly.

\paragraph{One-sided trend modulation.}
Exponential aggregation estimates the local level of teacher support, but it
does not explicitly distinguish a stable positive region from one whose
support is rapidly decreasing. To capture this direction, \methodname{} fits
an ordinary least-squares slope over the valid positions in the
$N_{\max}$ window. For a complete window of length $n$, the slope can be
written as
\begin{equation}
s_t
=
\frac{
\sum_{i=0}^{n-1}
(i-\bar{i})\delta_{t+i}
}{
\sum_{i=0}^{n-1}
(i-\bar{i})^2
},
\qquad
\bar{i}
=
\frac{n-1}{2}.
\label{eq:ols_slope_analysis}
\end{equation}
Since the slope coefficients sum to zero, adding a constant to all gaps does
not change $s_t$. The slope therefore captures local direction rather than
the absolute magnitude of teacher support. Equation~\eqref{eq:ols_slope_analysis}
omits the numerical stabilizer for clarity; the implementation adds
$\epsilon_{\mathrm{slope}}>0$ to the denominator and sets the slope to zero
when fewer than two valid tokens are available.

Under independent noise with variance $\sigma^2$, the slope variance is
\begin{equation}
\operatorname{Var}[s_t]
=
\frac{\sigma^2}
{\sum_{i=0}^{n-1}(i-\bar{i})^2}
=
\frac{12\sigma^2}
{n(n^2-1)}.
\label{eq:ols_slope_variance}
\end{equation}
For $n=8$, this gives $\operatorname{Var}[s_t]=\sigma^2/42$. Estimating the
trend across several positions is therefore less sensitive to a single noisy
gap than comparing only two adjacent tokens.

To make the slope comparable across responses with different gap scales, we
normalize it as
\begin{equation}
\widetilde{s}_t
=
\frac{s_t}
{
M^{-1}\sum_{j=1}^{M}|\delta_j|
+
\epsilon_{\mathrm{scale}}
},
\label{eq:normalized_slope_analysis}
\end{equation}
where $M$ is the number of valid response tokens and
$\epsilon_{\mathrm{scale}}>0$ prevents division by a near-zero scale. If all gaps are multiplied
by a positive constant, both the slope and the normalization scale change by
the same factor. Consequently, $\widetilde{s}_t$ is approximately invariant
to the overall magnitude of the response-level gaps.

The one-sided trend factor is
\begin{equation}
\eta_t
=
\operatorname{clip}
\left(
1-\gamma
\max(-\widetilde{s}_t,0),
0,
1
\right).
\label{eq:one_sided_trend_factor}
\end{equation}
This construction has three useful properties. First, if
$\widetilde{s}_t\geq0$, then $\eta_t=1$, so an increasing trend does not amplify
the base distillation weight. Positive teacher support is already reflected
by the local gap and sigmoid gate; rewarding it again through the trend term
could create redundant amplification. Second, if
$\widetilde{s}_t<0$, the factor decreases smoothly as teacher support falls.
Third,
\begin{equation}
0
\leq \eta_t\leq1,
\label{eq:trend_factor_bound}
\end{equation}
The final token weight is
\begin{equation}
w_t
=
\sigma\left(
\beta_{\mathrm{gate}}
\bar{\delta}_t^{\mathrm{ad}}
\right)\eta_t.
\label{eq:trend_modulated_weight}
\end{equation}
Therefore, the trend term can only attenuate the sigmoid gate and cannot
increase it. The parameter $\gamma$ controls how rapidly the factor decreases
with a negative normalized slope. With $\gamma=0.3$, a normalized slope of
$-1$ gives $\eta_t=0.7$ before the lower clipping bound is reached.

\begin{table}[h]
\centering

\small
\setlength{\tabcolsep}{3.5pt}
\begin{tabular}{@{}lccccc@{}}
\toprule
\textbf{Method} & $G$ & $\epsilon_{\mathrm{clip}}$ &
$\lambda_{\mathrm{distill}}$ & $\beta_{\mathrm{gate}}$ & Ret. \\
\midrule
GRPO~\cite{shao2024deepseekmath}              & 8  & 0.2 & --    & --  & -- \\
Skill-GRPO / Skill-GRPO*       & 8  & 0.2 & --    & --  & KM \\
OPSD~\cite{zhao2026self}        & -- & --  & 0.01  & 5.0 & KM \\
Skill-SD          & 8  & 0.2 & 0.001 & --  & KM \\
GRPO+OPSD         & 8  & 0.2 & 0.01  & 0.0 & KM \\
RLSD~\cite{yang2026self}              & 8  & 0.2 & 0.5   & --  & KM \\
SDAR~\cite{lu2026self}              & 8  & 0.2 & 0.01  & 5.0 & KM \\
\methodname{}     & 8  & 0.2 & 0.01  & 5.0 & KM \\
\bottomrule
\end{tabular}
\caption{Method-specific hyperparameter settings used in our experiments.
Ret. denotes the skill-retrieval strategy used during training.
The starred variant additionally receives retrieved skills at evaluation;
both Skill-GRPO variants otherwise use the same training configuration.
Prompt-only baselines are omitted because they have no optimization
hyperparameters.}
\label{tab:hyperparams}
\end{table}

\begin{table}[h]
\centering
{
\small
\setlength{\tabcolsep}{3.5pt}
\begin{tabular}{@{}lcl@{}}
\toprule
\textbf{Parameter} & \textbf{Value} & \textbf{Description} \\
\midrule
$N_{\min}$                  & 1    & Short aggregation window \\
$N_{\max}$                  & 8    & Long, variance, and trend window \\
$\alpha$                    & 0.8  & Exponential decay factor \\
$\tau_{\mathrm{low}}$       & 0.05 & Lower variance threshold \\
$\tau_{\mathrm{high}}$      & 0.5  & Upper variance threshold \\
$\gamma$                    & 0.3  & Negative-trend modulation strength \\
$\beta_{\mathrm{gate}}$     & 5.0  & Sigmoid gate sharpness \\
$\lambda_{\mathrm{PCSD}}$   & 0.01 & Distillation loss coefficient \\
\bottomrule
\end{tabular}
}
\caption{Hyperparameters specific to \methodname{}.}
\label{tab:pcsd_hyperparams}
\end{table}

\paragraph{Complementary roles.}
The exponentially aggregated gap and one-sided trend factor capture different
aspects of local consistency. The former estimates the level of teacher
support while reducing isolated fluctuations; the latter detects whether
that support is persistently decreasing. A token may have a positive
aggregated gap but a negative local slope, indicating that the teacher
advantage is weakening. In this case, the trend factor conservatively reduces
the associated distillation weight. In contrast, a nonnegative slope leaves
the base weight unchanged.

This asymmetric design prevents the trend estimator from creating additional
credit on its own. It acts only as a bounded attenuation mechanism, while the
magnitude of the distillation signal remains determined by the aggregated
teacher--student gap. Together, exponential aggregation and one-sided trend
modulation provide noise reduction, temporal locality, and protection against
stale or declining teacher support.

\begin{table*}[t]
\centering
\small
\setlength{\tabcolsep}{3.5pt}
\renewcommand{\arraystretch}{1.18}
\begin{tabular}{
@{}c
>{\raggedright\arraybackslash}p{0.29\textwidth}
>{\raggedright\arraybackslash}p{0.35\textwidth}
>{\raggedright\arraybackslash}p{0.22\textwidth}
@{}}
\toprule
\textbf{Step} & \textbf{Observation} &
\textbf{Reasoning Summary} & \textbf{Action} \\
\midrule
\multicolumn{4}{@{}l@{}}{\textbf{Task:}
\texttt{heat some potato and put it in garbagecan}} \\
\midrule
1 &
The agent starts in a room containing several cabinets, countertops, a
fridge, a microwave, and a garbage can. &
A potato is likely to be stored in the fridge, a cabinet, or on a
countertop; inspect the fridge first. &
\texttt{go to fridge 1} \\

2 &
The agent arrives at \texttt{fridge 1}, which is closed. &
Open the fridge to determine whether it contains the target object. &
\texttt{open fridge 1} \\

3 &
The opened fridge contains several objects, but no potato. &
Continue the search at the countertops and cabinets. &
\texttt{go to countertop 1} \\

4 &
\texttt{countertop 1} contains \texttt{potato 1} together with several
other objects. &
The target has been located and must be acquired before heating. &
\texttt{take potato 1 from countertop 1} \\

5 &
The agent successfully picks up \texttt{potato 1}. &
Navigate to the microwave to perform the required state transformation. &
\texttt{go to microwave 1} \\

6 &
The agent arrives at \texttt{microwave 1} while holding the potato. &
Use the microwave to heat the acquired potato. &
\texttt{heat potato 1 with microwave 1} \\

7 &
The environment confirms that \texttt{potato 1} has been heated. &
The remaining subgoal is to transport the heated potato to the target
receptacle. &
\texttt{go to garbagecan 1} \\

8 &
The agent arrives at \texttt{garbagecan 1} while still holding the heated
potato. &
Place the heated potato in the garbage can to complete the task. &
\texttt{move potato 1 to garbagecan 1} \\
\midrule
\multicolumn{3}{r}{\textbf{Final outcome:}} &
\texttt{won=True}, reward $=10.0$ \\
\bottomrule
\end{tabular}
\caption{A complete successful rollout on an unseen ALFWorld environment.
The agent recovers from an unsuccessful initial search in the fridge, locates
the potato on a countertop, preserves the required acquire--heat--place
ordering, and completes the task in eight valid actions.}
\label{tab:successful_rollout}
\end{table*}

\subsection{Training and Hyperparameter Settings}
Tables~\ref{tab:hyperparams} and~\ref{tab:pcsd_hyperparams} summarize the shared and method-specific settings. All trainable methods are optimized with AdamW using a learning rate of $1\times10^{-6}$, a KL penalty coefficient of $0.01$ toward the reference policy, gradient clipping with a maximum norm of $1.0$, and 150 update steps. On ALFWorld, each batch contains 16 tasks with eight rollouts per task, yielding 128 trajectories, and the maximum prompt length is 2,048 tokens. WebShop uses the same number of tasks and rollouts with a maximum prompt length of 4,096 tokens. The teacher is initialized from the same base checkpoint as the student and remains frozen throughout training. Teacher log-probabilities and token-level weights are detached during optimization. Keyword matching (KM) is used whenever skill retrieval is required, while evaluation-time skills are provided only to methods marked with *.

For \methodname{}, we use a pointwise short window $N_{\min}=1$ and a long window $N_{\max}=8$, with the latter also used to estimate local variance and the OLS trend. The exponential decay factor is set to $\alpha=0.8$. Local variance is mapped between the short- and long-window estimates using $\tau_{\mathrm{low}}=0.05$ and $\tau_{\mathrm{high}}=0.5$. We set the negative-trend modulation strength to $\gamma=0.3$, the sigmoid sharpness to $\beta_{\mathrm{gate}}=5.0$, and the distillation coefficient to $\lambda_{\mathrm{PCSD}}=0.01$. Unless otherwise specified, these settings are fixed across both backbones and benchmarks without task-specific tuning.

\begin{figure*}[t]
\centering
\includegraphics[width=0.9\textwidth]
{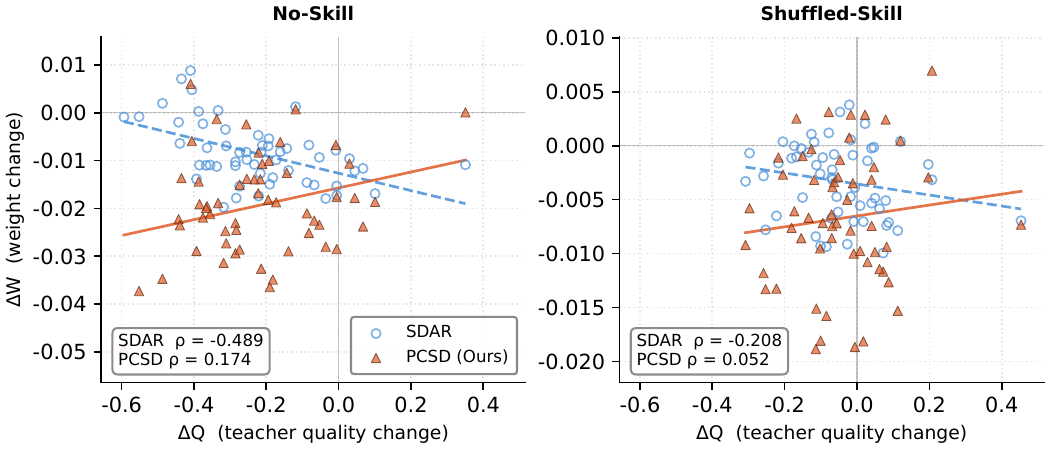}
\caption{Relationship between teacher-quality changes and weight
changes under skill-removal and shuffled-skill perturbations. Dashed
and solid lines show method-specific linear fits, while the annotations
report Spearman correlations. PCSD exhibits weak nonnegative
associations, whereas SDAR shows inverse associations under both
perturbations.}
\label{fig:quality_alignment}
\end{figure*}

\begin{figure}[h]
\centering
\includegraphics[width=\columnwidth]
{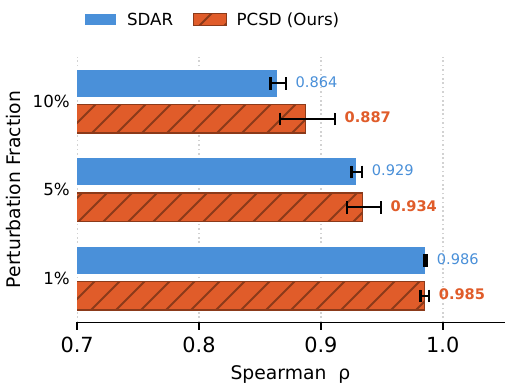}
\caption{Robustness of token-weight rankings under isolated-gap
perturbations. Spearman correlation is computed between the original
and perturbed token-weight rankings. PCSD remains comparable to SDAR
under light perturbation and yields higher mean rank preservation under
the $5\%$ and $10\%$ perturbations. Error bars indicate variability
across repeated perturbation draws.}
\label{fig:rank_preservation}
\end{figure}

\subsection{Benchmark Protocols}
\label{app:benchmark_protocols}

The batch and evaluation sizes below refer to the number of tasks or episodes processed in one training update or evaluation run, rather than the total number of instances in each benchmark split.

\paragraph{ALFWorld.}
Following the GiGPO setup used in the main paper, we train on the
official \texttt{train} split and evaluate in-distribution performance
on \texttt{valid\_seen}. Generalization is evaluated separately on
\texttt{valid\_unseen}, which contains unseen environments and layouts.
Each training update samples 16 tasks with eight rollouts per task,
yielding 128 trajectories, while each periodic validation run evaluates
128 episodes from \texttt{valid\_seen}.

Each episode is limited to 50 environment actions. The maximum prompt and response lengths are 2,048 and 512 tokens, respectively. Episode success is determined by the final binary \texttt{won} signal returned by the environment. Successful trajectories receive a training reward of 10, whereas failures, invalid terminations, and timeouts receive 0.
This reward scaling is used only for training and does not affect the
reported success rate:
\begin{equation}
\mathrm{SR}
=
\frac{100}{N_{\mathrm{eval}}}
\sum_{n=1}^{N_{\mathrm{eval}}}\mathrm{won}_n,
\qquad
\mathrm{won}_n\in\{0,1\}.
\label{eq:alfworld_sr}
\end{equation}

\paragraph{WebShop.}
We use 1,000 training tasks and evaluate on 128 fixed validation instances. The training and validation environments are initialized with different fixed random seeds, using $s$ for training and $s+1000$ for validation. Each training update processes 16 tasks with eight rollouts per task, yielding 128 trajectories. Each episode is limited to 15 interaction steps, with maximum prompt and response lengths of 4,096 and 512 tokens, respectively. Evaluation uses one sampled trajectory per instance with temperature $0.4$ and a fixed evaluation seed.

WebShop returns a continuous task score measuring product--specification alignment. Let $q_n\in[0,1]$ denote the task score of episode $n$, and let $z_n\in\{0,1\}$ indicate whether the episode terminates with $q_n=1$. Successful trajectories receive a training reward of 10, while all other trajectories receive 0. We report the normalized mean task score and strict task success rate:
\begin{equation}
\begin{aligned}
\mathrm{Score}
&=\frac{100}{N}\sum_{n=1}^{N}q_n,\\
\mathrm{Acc}
&=\frac{100}{N}\sum_{n=1}^{N}z_n.
\end{aligned}
\label{eq:webshop_metrics}
\end{equation}
Score captures partial product--specification satisfaction, whereas Acc measures the proportion of fully completed tasks.

\subsection{Qualitative Rollout}
\label{app:qualitative_rollout}

Table~\ref{tab:successful_rollout} presents a complete successful PCSD trajectory. This trajectory was generated by the PCSD-trained checkpoint under the \texttt{valid\_unseen} evaluation protocol. Observations and reasoning are condensed for readability, while the actions and terminal outcome are transcribed from the recorded rollout. The task requires the agent to
locate a potato, acquire it, heat it, and place it in the garbage can. The
trajectory contains eight valid environment actions and terminates with
\texttt{won=True} and a reward of 10. The reasoning is abbreviated for
readability, while the observations and actions preserve the recorded
interaction sequence.

\subsection{Weight Robustness and Teacher-Quality Alignment}
\label{app:weight_diagnostics}

\paragraph{Robustness of weight rankings.}
We first examine whether sparse, isolated perturbations disrupt the
global ordering of token-level distillation weights. We inject spikes
of magnitude $3.0$ into randomly selected teacher--student gap values,
using perturbation fractions of $1\%$, $5\%$, and $10\%$. We then
compute the Spearman correlation between the original and perturbed
token-weight rankings. The same gap sequences and perturbation
locations are used for both methods.

As shown in Figure~\ref{fig:rank_preservation}, PCSD and SDAR perform
similarly under the $1\%$ perturbation, with correlations of $0.985$
and $0.986$, respectively. Under stronger perturbations, PCSD yields
higher mean correlations: $0.934$ versus $0.929$ at $5\%$, and $0.887$
versus $0.864$ at $10\%$. These results suggest that PCSD better
preserves the relative priority of tokens under moderate and severe
isolated-gap noise, while the difference under light perturbation is
negligible.

\paragraph{Alignment with teacher-quality changes.}
We next examine whether changes in token weights follow changes in
teacher quality. Let $Q$ denote the teacher confidence assigned to an
expert action and $W$ the corresponding token-level distillation
weight. For each evaluated state--action pair, we compute
$\Delta Q=Q_{\mathrm{perturbed}}-Q_{\mathrm{original}}$ and
$\Delta W=W_{\mathrm{perturbed}}-W_{\mathrm{original}}$.
We consider two perturbations: removing the retrieved skill context
and replacing it with a shuffled skill. A directionally consistent
weighting rule should produce a nonnegative association between
$\Delta Q$ and $\Delta W$, because reduced teacher confidence should
generally be accompanied by reduced distillation weight.

Figure~\ref{fig:quality_alignment} shows that PCSD has weak but
nonnegative Spearman correlations under both perturbations:
$\rho=0.174$ for skill removal and $\rho=0.052$ for shuffled skills.
In contrast, SDAR produces negative correlations of $-0.489$ and
$-0.208$, respectively. These results do not indicate strong
teacher-quality alignment for every sample, particularly under
shuffled skills. However, they show that PCSD avoids the systematic
inverse association observed for SDAR. This behavior is consistent
with PCSD's independently computed sigmoid weights, whereas normalized
weighting may introduce competition across tokens.

Together, these diagnostics provide complementary evidence that PCSD
preserves token priorities under isolated noise and avoids the inverse
association observed for SDAR under teacher-quality perturbations. They should be
interpreted as analyses of the weighting behavior rather than direct
evidence of downstream task performance or causal attribution to an
individual component.


\end{document}